%% file: iclr2027_conference.tex
\documentclass{article}
\PassOptionsToPackage{table}{xcolor}

\newcommand{\paperpath}{}
\makeatletter
\IfFileExists{iclr2027_conference.sty}{}{%
  \renewcommand{\paperpath}{iclr2027/}%
  \def\input@path{{iclr2027/}}
}
\makeatother
\usepackage{iclr2027_conference,times}
\input{\paperpath preamble}
\graphicspath{{\paperpath}}
\usepackage{tikz}
\definecolor{beataccent}{HTML}{A33A46}

\DeclareRobustCommand{\beatclappernotemark}{%
  \begin{tikzpicture}[x=0.032em,y=-0.032em,baseline=-0.80em]
    \path[use as bounding box] (0,0) rectangle (28,28);
    \draw[beataccent,line width=0.065em,rounded corners=0.045em]
      (3,12) rectangle (25,26);
    \fill[beataccent] (2,7) -- (24,2) -- (25,6) -- (3,11) -- cycle;
    \fill[white] (7,5.86) -- (10,5.18) -- (8,9.86) -- (5,10.55) -- cycle;
    \fill[white] (15,4.05) -- (18,3.36) -- (16,8.05) -- (13,8.73) -- cycle;
    \fill[beataccent] (14,15) rectangle (16,22.5);
    \fill[beataccent] (12.5,22.5) ellipse[x radius=3.5,y radius=2.2];
    \fill[beataccent] (16,15) .. controls (18,16) and (21,16) .. (20,20)
      .. controls (19,18) and (17,18) .. (16,18) -- cycle;
  \end{tikzpicture}%
}

\title{\beatclappernotemark\hspace{0.18em}BEAT: Rhythm-Elastic Alignment for Agentic Music-guided Movie Trailer Generation}

\author{%
  Yutong Wang\textsuperscript{1},
  Yunke Wang\textsuperscript{1},
  \textbf{Xinyuan Chen\textsuperscript{2}\footnotemark[1]} \ ,
  \textbf{Chang Xu\textsuperscript{1}\thanks{Corresponding author.}}\\
  {\textsuperscript{1}The University of Sydney\quad
    \textsuperscript{2}Shanghai AI Laboratory
  }\\
}

\iclrfinalcopy 
\begin{document}
\maketitle

\input{sections/abstract}
\input{sections/introduction}
\input{sections/related_works}

\input{sections/method}
\input{sections/benchmark}
\input{sections/experiments}
\input{sections/conclusion}

\input{sections/08_statements}
\bibliographystyle{iclr2027_conference}
\bibliography{refs}

\clearpage
\input{sections/appendix}

\end{document}

%% file: preamble.tex
\usepackage[utf8]{inputenc} 
\usepackage[T1]{fontenc}    
\usepackage{hyperref}       
\usepackage{url}            
\usepackage{booktabs}       
\usepackage{amsfonts}       
\usepackage{amsmath}        
\usepackage{amssymb}        
\usepackage{nicefrac}       
\usepackage{microtype}      
\usepackage[table]{xcolor}
\usepackage{graphicx}       
\usepackage{algorithm}      
\usepackage{algorithmic}    
\usepackage{multirow}       
\usepackage{subcaption}     
\usepackage{wrapfig}        
\usepackage{tikz}
\usepackage{pifont}
\newcommand{\cmark}{\makebox[1.2em][c]{\ding{51}}}%
\newcommand{\xmark}{\makebox[1.2em][c]{\ding{55}}}%
\usepackage{bm}
\usepackage{xspace}

\DeclareMathOperator*{\argmax}{arg\,max}

\DeclareMathOperator{\softmax}{softmax}
\DeclareMathOperator{\cosim}{cos}
\DeclareMathOperator{\KL}{KL}

\newcommand{\R}{\mathbb{R}}

\newcommand{\ba}{\mathbf{a}}
\newcommand{\bv}{\mathbf{v}}

\newcommand{\bS}{\mathbf{S}}
\newcommand{\bpi}{\boldsymbol{\pi}}
\newcommand{\cM}{\mathcal{M}}

\newcommand{\cS}{\mathcal{S}}
\newcommand{\cB}{\mathcal{B}}
\newcommand{\cL}{\mathcal{L}}

\definecolor{metrics2}{HTML}{e5e5e5}
\definecolor{metrics1}{HTML}{F0D695}
\definecolor{metrics3}{HTML}{82ACD1}
\definecolor{metrics4}{HTML}{98B567}
\definecolor{tabhighlight}{HTML}{e5e5e5}

%% file: sections/abstract.tex
\begin{abstract}
Automatic movie trailer generation selects shots from a full-length film and synchronizes them with chosen music.
Existing methods either relegate music alignment to post-processing or enforce rigid one-to-one shot-music mappings, overlooking that professional editing rhythm is \emph{elastic}: 
energetic passages favor rapid cuts, while quieter passages allow shots to span multiple bars.
We introduce \textbf{BEAT}, a framework that addresses this gap with two core components: MuVA, a compact music-visual alignment encoder trained with Sinkhorn-regularized two-stage learning, and Bar-DP, an energy-adaptive dynamic programming algorithm that produces elastic many-to-one alignments.
These components are integrated into a five-phase agentic pipeline that grounds the core alignment in learned cross-modal features while coordinating higher-level creative decisions through structured text signals.
To support comprehensive evaluation, we also introduce TrailerArena, a benchmark with 20+ metrics across four complementary dimensions.
On TrailerArena, BEAT achieves state-of-the-art performance across shot selection, ordering, and perceptual quality, while producing fully composed trailers end-to-end. 
Code and models are available at \url{https://github.com/hhhh1138/BEAT}.
\end{abstract}

%% file: sections/introduction.tex
\section{Introduction}

Movie trailers condense a film into a short engaging audiovisual sequence while avoiding key plot revelations.
A chosen soundtrack provides emotional direction and rhythmic structure, motivating \emph{music-guided trailer generation}~\citep{xu2015trailer,wang2024inverse}: selecting and arranging shots from a given movie to fit the music.
This requires coherent shot selection and \emph{elastic} timing: energetic passages favor rapid cuts, while quieter passages allow a shot to span multiple bars.

Learning-based approaches provide useful mechanisms for shot selection, but shot compatibility and editing duration remain distinct decisions.
IPOT~\citep{wang2024inverse} and MMSC~\citep{zhu2025weakly} model shot-music matching through optimal transport or cross-attention, while SSMP~\citep{zhu2025self} predicts shot sequences and handles music alignment in post-processing.
A rigid one-to-one matching does not express how a shot can extend across several bars as musical energy changes.
Trailer generation therefore calls for an alignment mechanism that couples learned shot-music compatibility with variable shot durations.

Agentic systems offer complementary capabilities for planning and iterative editing.
TRAILDREAMS~\citep{balestri2025traildreams} coordinates trailer creation through LLM-driven sequence selection and audio generation; DIRECT~\citep{li2026direct} organizes video mashup creation through hierarchical planning; and CutClaw~\citep{cutclaw2025} coordinates agents for music-synchronized editing of long-form footage.
However, repeated model calls across planning and editing can incur substantial inference costs.
When editing decisions depend on model reasoning and prompt design, rhythmic constraints become difficult to specify and verify explicitly.
This motivates a framework that combines agent-based planning with a trainable alignment model and explicit control over shot durations.

Our key idea is to learn bar-shot compatibility, select shot durations according to musical energy, and coordinate creative decisions through agents.
We introduce \textbf{BEAT} (\textbf{B}ar-level \textbf{E}lastic \textbf{A}lignment for agentic \textbf{T}railer generation), which implements this idea through a compact alignment model and elastic selection within a five-phase agentic pipeline.
MuVA scores candidate shots for each musical bar; Bar-DP uses these scores and musical energy to select shots and their bar spans.
LLM/VLM agents coordinate planning, composition, and iterative critique through structured text signals.
Given a movie and a music, BEAT produces a fully composed trailer in a single run.
Our contributions are:

\begin{itemize}
    \item \textbf{Learned alignment with elastic timing.} We introduce MuVA, a compact cross-modal encoder for bar-shot compatibility, and Bar-DP, an energy-adaptive selection algorithm that jointly determines shot choice and duration. Their many-to-one alignments support rapid cuts and sustained shots spanning multiple bars within a unified formulation (\S\ref{sec:muva}, \S\ref{sec:bardp}).

    \item \textbf{Agentic trailer generation.} We integrate learned alignment and elastic selection into a five-phase workflow from music analysis to final composition. Structured text signals connect user instructions, narrative planning, and tagline generation with iterative critique, allowing agent feedback to guide re-selection while retaining explicit control over shot spans (\S\ref{sec:pipeline}).

    \item \textbf{Comprehensive trailer evaluation.} We introduce TrailerArena to assess selection, ordering, composition, and perceptual quality, including Fr\'{e}chet Shot Distance for comparing selected-shot feature distributions. BEAT leads the compared methods in F1, SoftF1@5, and FSD on both Test-8 and Test-73, with ablations examining component contributions (\S\ref{sec:benchmark}, \S\ref{sec:experiments}).
\end{itemize}

%% file: sections/related_works.tex
\section{Related Work}

\paragraph{Movie Trailer Generation.}
Early approaches used hand-crafted heuristics on multimodal features~\citep{smith2017harnessing,smeaton2006automatically,hesham2018smart}, graph-based narrative analysis to detect turning points~\citep{papalampidi2021film}, or cinematic grammar to extract logos, dialogue, and action scenes~\citep{chen2004action,liu2015semi,JCST-2212-13064}.
More recent methods learn trailer composition from paired movie-trailer data.
TGT~\citep{atapattu2024towards} models trailer generation as sequence-to-sequence translation with an autoregressive decoder, while IPOT~\citep{wang2024inverse} jointly optimizes shot selection and music alignment through inverse partial optimal transport.
MMSC~\citep{zhu2025weakly} learns weakly supervised trailerness scores through multimodal cross-attention with semantic consistency constraints.
SSMP~\citep{zhu2025self} trains a Transformer encoder for masked prediction with self-corrective re-masking.
Despite these advances, rigid one-to-one shot-music correspondences do not capture the elastic editing rhythm of professional trailers.

\paragraph{Multi-Agent Systems for Video Production.}
Agentic workflows coordinate video editing~\citep{wang2024lave,li2025shots,ding2025prompt,zhou2026videoagent} and film production~\citep{wu2025automated,chen2025filmagent,balestri2025traildreams} through specialized roles and tools.
EditDuet~\citep{sandoval2025editduet} couples an editor with a critic, while DIRECT~\citep{li2026direct} organizes video mashup creation into a Screenwriter--Director--Editor hierarchy.
CineAgents~\citep{zhang2025cinebench} combines hierarchical narrative memory with iterative planning for instruction-driven cinematic compilation.
Closer to music-guided editing, CutClaw~\citep{cutclaw2025} coordinates narrative planning, editing, and review to synchronize selected footage with music.
BEAT combines agent coordination with a trained bar-shot alignment model and energy-adaptive elastic selection, using structured text signals to convey instructions, taglines, and critique.

%% file: sections/method.tex
\begin{figure}[t]
    \centering
    \includegraphics[width=\linewidth]{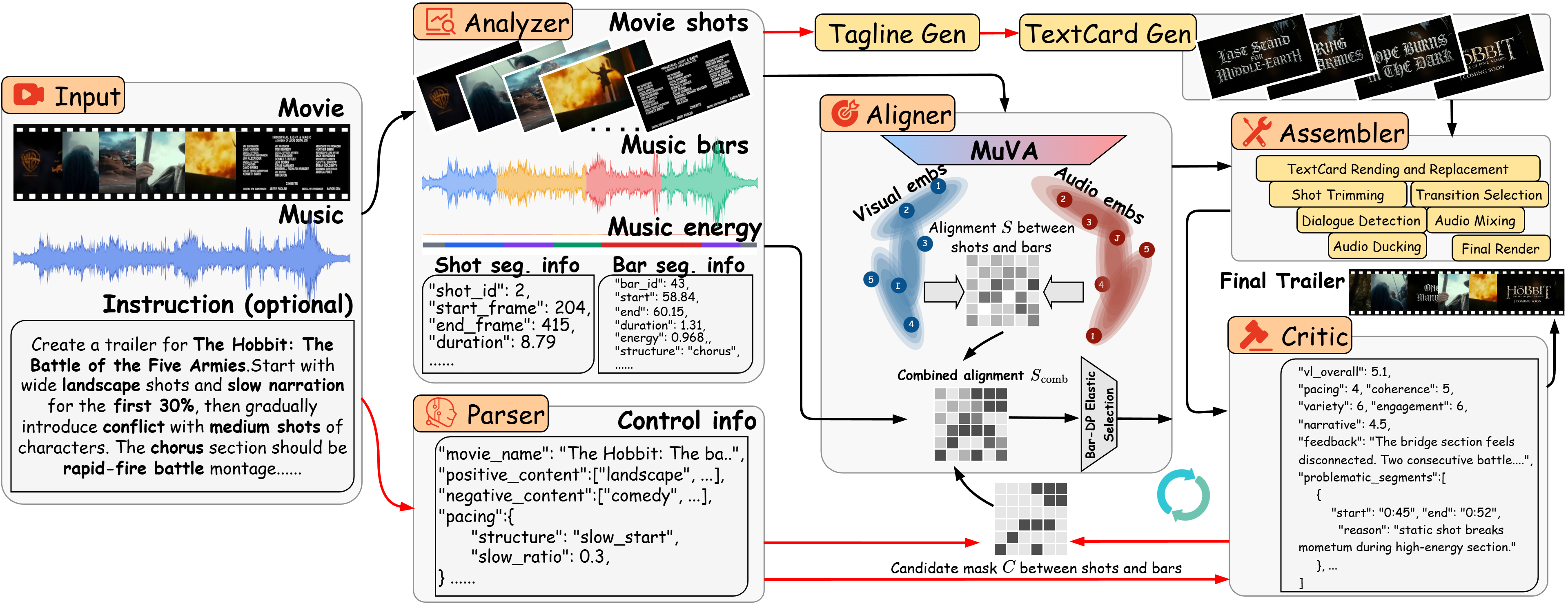}
    \caption{\textbf{Overview of BEAT.} 
    Given a movie, trailer music, and an optional text instruction, the pipeline operates in five phases. 
    The \textbf{Analyzer} segments the movie into shots and the music into bars, extracting visual/audio features and per-bar energy estimates. 
    An optional \textbf{Parser} converts free-form instructions into a structured control information that modifies the candidate mask. 
    The \textbf{Aligner} computes cross-modal alignment scores, which are fused with energy-dynamics matching and fed to Bar-DP for elastic shot selection. 
    The \textbf{Assembler} generates taglines, renders styled text cards, applies transitions and dialogue ducking, and produces the final trailer. 
    The \textbf{Critic} reviews the rendered trailer and returns timestamped feedback to trigger re-selection if needed. 
    Structured text signals (red arrows) coordinate modules across phases.}
    \label{fig:pipeline}
\end{figure}

\section{Proposed Method}
\label{sec:method}

We present BEAT, a framework for music-guided movie trailer generation.
Given a movie and a piece of trailer music, BEAT selects and arranges movie shots to produce a trailer that is rhythmically synchronized with the music, narratively coherent, and spoiler-free.
We first formalize the problem (\S\ref{sec:formulation}), then describe the three core components: the Music-Visual Aligner (MuVA, \S\ref{sec:muva}), the elastic Bar-DP selection algorithm (\S\ref{sec:bardp}), and the agentic generation pipeline (\S\ref{sec:pipeline}).

\subsection{Problem Formulation}
\label{sec:formulation}

Let $\cM = \{s_1, s_2, \ldots, s_I\}$ denote the set of $I$ shots obtained by segmenting a full-length movie via a shot boundary detector~\citep{soucek2024transnet}.
Each shot $s_i$ is characterized by a visual feature vector $\bv_i \in \R^{d_v}$ extracted by a pretrained visual encoder, and a temporal duration $\delta_i > 0$.

The trailer music is segmented into $J$ musical bars $\cB = \{b_1, b_2, \ldots, b_J\}$ via change-point detection on the audio signal~\citep{truong2020selective}.
Each bar $b_j$ is represented by an audio feature vector $\ba_j \in \R^{d_a}$ extracted by a pretrained audio encoder, a duration $\ell_j > 0$, and an energy level $e_j \in [0, 1]$ computed as the normalized RMS amplitude.

\paragraph{Elastic Alignment.}
We define a trailer as an \emph{elastic alignment} $\bpi$ that maps the music bar sequence to a shot assignment sequence.
Formally, $\bpi = \bigl((c_1, r_1), \ldots, (c_K, r_K)\bigr)$, where segment $k$ assigns shot $c_k \in \{1, \ldots, I\}$ to bars $[r_k, r_{k+1})$, with $r_1 = 1$, $r_{K+1} = J{+}1$.
The number of segments $K \leq J$ is determined by the algorithm, and a single shot may span multiple consecutive bars (many-to-one mapping), while each bar belongs to exactly one shot.
This elastic formulation generalizes the rigid one-to-one mapping used in prior works~\citep{wang2024inverse,zhu2025self,zhu2025weakly,xu2015trailer}.

\paragraph{Objective.}
The goal is to find an elastic alignment $\bpi^*$ that maximizes a composite quality score:
\begin{equation}
\label{eq:objective}
\bpi^* = \argmax_{\bpi \in \Pi} \sum_{k=1}^{K} \Biggl[
    \underbrace{\bar{f}(c_k, r_k, r_{k+1})}_{\text{alignment}}
    + \lambda_{\text{sm}} \cdot \underbrace{g(c_{k-1}, c_k)}_{\text{smoothness}}
    + \lambda_{\text{cut}} \cdot \underbrace{h(e_{r_k:r_{k+1}})}_{\text{cut bonus}}
\Biggr],
\end{equation}
subject to no-repeat constraints $c_k \neq c_{k'}$ for $k \neq k'$ and duration constraints $\sum_{j=r_k}^{r_{k+1}-1} \ell_j \leq \delta_{c_k}/\eta$ for spans of multiple bars.
Here, $\bar{f}$ measures the average alignment quality between a shot and its assigned bars, $g$ penalizes extreme visual discontinuities between consecutive shots (without enforcing similarity, since trailers often use deliberate contrast), and $h$ provides an energy-adaptive cut bonus.
$\Pi$ denotes the space of valid elastic alignments.

\subsection{MuVA: Music-Visual Aligner}
\label{sec:muva}

MuVA is the core alignment model that scores the compatibility between each musical bar $b_j$ and each movie shot $s_i$.
It produces an alignment score matrix $\bS \in \R^{J \times I}$ where $\bS_{ji}$ indicates how well shot $s_i$ matches bar $b_j$.

\subsubsection{Architecture}

MuVA takes as input the raw audio features $\ba_j \in \R^{d_a}$ extracted by CLAP~\citep{wu2023large} ($d_a{=}512$) and visual features $\bv_i \in \R^{d_v}$ extracted by ImageBind~\citep{girdhar2023imagebind} ($d_v{=}1024$).
Linear projections with RMSNorm map both modalities into a shared $d$-dimensional space, yielding $\tilde{\ba}_j, \tilde{\bv}_i \in \R^d$.
Each modality is then processed independently by $L$ Transformer blocks with Rotary Position Embeddings (RoPE) to capture temporal ordering, producing self-attended representations $\bar{\ba}_j$ and $\bar{\bv}_i$.
The two streams interact through $L_{\text{ca}}$ layers of bidirectional cross-attention (audio attends to visual, then visual attends to the updated audio), yielding the final cross-modal representations $\hat{\ba}_j$ and $\hat{\bv}_i$.
The alignment score between bar $j$ and shot $i$ is then computed via temperature-scaled cosine similarity:
\begin{equation}
\label{eq:alignment_score}
\bS_{ji} = \frac{\langle \hat{\ba}_j, \hat{\bv}_i \rangle}{\|\hat{\ba}_j\| \cdot \|\hat{\bv}_i\|} \cdot \frac{1}{\tau},
\end{equation}
where $\tau$ is a learnable temperature parameter.

\subsubsection{Two-Stage Training}
\label{sec:training}

MuVA is trained in two stages to leverage both abundant video-music data and scarce trailer-specific annotations.

\paragraph{Stage 1: Pretraining on Video-Music Data.}
We pretrain on large video-music pairs from V2M~\citep{tian2025vidmuse}.
Since these pairs lack bar-shot alignment labels, we construct soft targets based on temporal IoU between music bars and visual shots.
Bar-shot pairs with greater temporal overlap receive higher alignment scores, while non-overlapping pairs are assigned zero.
The pretraining loss combines KL divergence on alignment distributions with an InfoNCE~\citep{oord2018infonce} contrastive term on pooled representations:
\begin{equation}
\label{eq:pretrain_loss}
\cL_{\text{pretrain}} = \KL\bigl(\softmax(\bS^{\text{soft}} / \tau_t) \;\|\; \softmax(\bS / \tau_t)\bigr) + \lambda_{\text{con}} \cdot \cL_{\text{InfoNCE}}.
\end{equation}

\paragraph{Stage 2: Fine-tuning on Trailer Alignments.}
We fine-tune on trailer-movie pairs with binary bar-to-shot annotations $\bS^{\text{gt}}$, allowing multiple positive shots per bar.
For each annotated bar, we normalize the target as $q_{ji}=\bS^{\text{gt}}_{ji}/\sum_{i'}\bS^{\text{gt}}_{ji'}$ and define $\mathbf{p}_j=\softmax(\bS_{j,:}/\tau_t)$ over candidate shots.
KL alignment supervises per-bar correctness, but without structural constraints, a few visually distinctive shots can dominate scores across many bars.
We add Sinkhorn regularization~\citep{cuturi2013sinkhorn}, which iteratively projects $\bS$ toward a doubly-stochastic matrix, enforcing that each bar focuses on few shots and each shot is not claimed by too many bars:
\begin{equation}
\label{eq:sinkhorn}
\bS^{(0)} = \exp(\bS / \tau_s), \quad
\bS^{(t+1)} = \mathcal{N}_c\bigl(\mathcal{N}_r(\bS^{(t)})\bigr), \quad t = 0, \ldots, T{-}1,
\end{equation}
where $\mathcal{N}_r$ and $\mathcal{N}_c$ normalize rows and columns respectively, and $\tau_s$ controls the sharpness of the target.
The regularization loss $\cL_{\text{sink}}$ encourages the model's raw scores to approximate this structured target.
We use $T{=}3$ iterations, applying a gentle structural prior; heavier settings ($T \geq 10$) over-constrain the model and degrade performance.
The full fine-tuning loss is:
\begin{equation}
\label{eq:full_loss}
\cL_{\text{ft}} =
  \underbrace{\frac{1}{J}\sum_{j=1}^{J}\KL\bigl(\mathbf{q}_j \;\|\; \mathbf{p}_j\bigr)}_{\cL_{\text{KL}} \text{ for selecting correctly}}
  + \lambda_{\text{sink}} \cdot \underbrace{\KL\bigl(\bS^{(T)} \;\|\; \softmax(\bS)\bigr)}_{\cL_{\text{sink}} \text{ for distributing evenly}}
  + \lambda_{\text{con}} \cdot \underbrace{\cL_{\text{InfoNCE}}}_{\substack{\cL_{\text{InfoNCE}} \text{ for globally}\\\text{aligning modalities}}},
\end{equation}
Bars without positive annotations contribute zero to $\cL_{\text{KL}}$; losses are averaged over examples.

\subsection{Bar-DP: Energy-Adaptive Elastic Selection}
\label{sec:bardp}

Given the alignment score matrix $\bS \in \R^{J \times I}$ from MuVA, Bar-DP approximates the optimum in Eq.~\eqref{eq:objective} via beam search over the bar sequence.
The key insight is that editing pace should adapt to musical dynamics: high-energy passages warrant rapid one-bar-one-shot cuts, while low-energy passages benefit from sustained shots spanning multiple bars.

\paragraph{State Representation.}
A state at bar position $j$ is a tuple $\sigma = (\text{score}, \text{assignments}, \text{used}, \text{last})$.
Here $\text{score} \in \R$ is the accumulated quality, $\text{assignments}$ records shot-bar span allocations, $\text{used} \subseteq \{1, \ldots, I\}$ tracks assigned shots and their visually similar neighbors, and $\text{last}$ is the most recently assigned shot.
We maintain a separate beam $\mathcal{H}_j$ for states whose next unassigned bar is $j$; assigning a span of $k$ bars advances a state to $\mathcal{H}_{j+k}$.

\paragraph{Transition Scoring.}
At each bar $j$, we consider assigning one of the top-$M$ candidate shots from $\bS_{j,:}$ with a bar span of $k \in \{1, \ldots, k_{\max}\}$:
\begin{equation}
\label{eq:bardp_score}
\text{score}(i, j, k) = \underbrace{\frac{1}{k}\sum_{j'=j}^{j+k-1} \bS_{j'i}}_{\text{avg.\ alignment}} + \lambda_{\text{sm}} \cdot \underbrace{\cosim(\bv_{\text{last}}, \bv_i)}_{\text{smoothness}} + \underbrace{\phi(e_{j:j+k})}_{\text{adaptive cut bonus}},
\end{equation}
where $\cosim(\cdot, \cdot)$ denotes cosine similarity and the energy-adaptive cut bonus $\phi$ is defined as:
\begin{equation}
\label{eq:cut_bonus}
\phi(e_{j:j+k}) = \lambda_{\text{cut}} \cdot \left(\bar{e}_{j:j+k} - 0.15\right),
\end{equation}
\noindent
\begin{minipage}[t]{0.48\textwidth}
with $\bar{e}_{j:j+k} = \frac{1}{k}\sum_{j'=j}^{j+k-1} e_{j'}$ being the average energy of the bar span.
When $\bar{e} = 1$ (high energy), the bonus is $+\lambda_{\text{cut}} \cdot 0.85$, encouraging cuts.
When $\bar{e} = 0$ (low energy), the penalty is $-\lambda_{\text{cut}} \cdot 0.15$, discouraging unnecessary transitions.

\paragraph{Duration Constraint.}
A shot $s_i$ may span $k > 1$ bars only if its duration $\delta_i$ can accommodate the total bar span:
\begin{equation}
\label{eq:duration_constraint}
\sum_{j'=j}^{j+k-1} \ell_{j'} \leq \frac{\delta_i}{\eta},
\end{equation}
where $\eta$ controls the permitted playback slowdown.

\paragraph{Neighbor Exclusion.}
To prevent redundant selections, we exclude previously used shots and their visually similar neighbors:
\begin{equation}
\text{used}' = \text{used} \cup \{i' : \cosim(\bv_i, \bv_{i'}) > \theta_{\text{sim}}\},
\end{equation}
where $\theta_{\text{sim}}$ is the similarity threshold.
We denote this exclusion set, including $i$ itself, by $\mathcal{N}(i)$.

\end{minipage}%
\hfill
\begin{minipage}[t]{0.48\textwidth}
\vspace{-0.5cm}
\begin{algorithm}[H]

\small
\caption{Bar-DP Elastic Selection}
\label{alg:bardp}
\begin{algorithmic}
\REQUIRE $\bS^{\text{comb}}$, $\{\ell_j\},\{\delta_i\},\{e_j\}$, $\mathcal{N}$
\ENSURE Approximate elastic alignment $\hat{\bpi}$
\STATE $\mathcal{H}_1 \!\leftarrow\! \{(0, [\,], \emptyset, \texttt{null})\}$
\STATE $\mathcal{H}_2,\ldots,\mathcal{H}_{J+1} \leftarrow \emptyset$
\FOR{$j = 1$ \TO $J$}
  \STATE $\mathcal{H}_j \leftarrow \text{Top-}W(\mathcal{H}_j)$
  \FOR{$(\textit{sc}, \textit{asgn}, \mathcal{U}, \textit{prev}) \in \mathcal{H}_j$}
    \FOR{$i \in \text{Top-}M(\bS^{\text{comb}}_{j,:}) \setminus \mathcal{U}$}
      \FOR{$k = 1$ \TO $\min(k_{\max},J-j+1)$}
        \IF{$\bS^{\text{comb}}_{j+k-1,i}=-\infty$ \OR ($k\!>\!1$
          \AND $\sum_{j'=j}^{j+k-1}\!\ell_{j'} \!>\! \delta_i/\eta$)}
          \STATE \textbf{break}
        \ENDIF
        \STATE $\Delta \!\leftarrow\! \text{score}(i,j,k)$
        \STATE $\mathcal{H}_{j+k} \!\leftarrow\! \mathcal{H}_{j+k} \!\cup\! \{(\textit{sc}{+}\Delta,$
        \STATE \quad $\textit{asgn}\mathbin{\Vert}(i,j,k), \mathcal{U}{\cup}\mathcal{N}(i), i)\}$
      \ENDFOR
    \ENDFOR
  \ENDFOR
\ENDFOR
\STATE \textbf{if} $\mathcal{H}_{J+1}=\emptyset$ \textbf{then return} failure
\STATE $\sigma^* \leftarrow \argmax_{\sigma \in \mathcal{H}_{J+1}} \sigma.\textit{sc}$
\RETURN alignment encoded by $\sigma^*.\textit{asgn}$
\end{algorithmic}
\end{algorithm}
\end{minipage}

Algorithm~\ref{alg:bardp} expands the top-$W$ states per bar using finite-score candidates and valid spans.
Transitions follow Eq.~\eqref{eq:bardp_score} with $\bS^{\text{comb}}$, $\text{last}=\textit{prev}$, and zero initial smoothness; $\Vert$ appends an assignment.
Each selected span $k_t^*=r_{t+1}-r_t$ is recovered from the best complete path; an empty $\mathcal{H}_{J+1}$ means no feasible completion was found.
Hyperparameters are listed in Appendix~\ref{app:bardp-settings}.

\paragraph{Complexity Analysis.}
Expanding $W$ states over $J$ bars, $M$ candidate shots, and up to $k_{\max}$ spans costs $O(J \cdot W \cdot M \cdot k_{\max})$.
Precomputing $\bS \in \R^{J \times I}$ and shot similarities for neighbor exclusion costs $O(J \cdot I \cdot d)$ and $O(I^2 \cdot d_v)$, respectively.
With the settings in Appendix~\ref{app:bardp-settings}, selection takes under 1.0\,s on a single CPU core, negligible compared to feature extraction and video rendering.

\subsection{Agentic Trailer Generation Pipeline}
\label{sec:pipeline}

Professional trailer editors do not produce a final cut in a single pass: they analyze the music, survey the footage, draft a rough cut, review it, and revise repeatedly.
BEAT mirrors this workflow through five phases connected by structured text signals, as shown in Figure~\ref{fig:pipeline}.
Unlike black-box agentic systems~\citep{cutclaw2025}, where all reasoning is delegated to LLMs, BEAT grounds the core alignment in a trained model.
LLM-generated text coordinates higher-level decisions: user instructions flow as keywords into the candidate mask, VLM taglines flow into composition, and critic feedback flows back to re-selection.

\paragraph{Phase 1--2: Analysis and Candidate Preparation.}
\textbf{MusicAnalyzer} segments the music into bars via change-point detection~\citep{truong2020selective}, extracts CLAP features~\citep{wu2023large}, and estimates per-bar energy and musical structure.
\textbf{NarrativeGuard} filters the shot pool by excluding spoiler-prone regions (final 15\%, opening logos) and scoring shot importance based on visual distinctiveness and position, producing a safe mask $\mathbf{m} \in [0,1]^I$.
The safe mask is broadcast to form a candidate mask $\bm{C} \in \{0,1\}^{J \times I}$.
When a user provides a free-form text instruction (e.g., ``focus on action scenes''), an LLM parses it into structured control signals, and ImageBind embeds the keywords into visual feature space to refine $\bm{C}$.

\paragraph{Phase 3: Alignment and Selection.}
MuVA computes the alignment matrix $\bS$ (\S\ref{sec:muva}), which is fused with energy-dynamics matching and shot importance:
\begin{equation}
\label{eq:score_combine}
\bS^{\text{comb}}_{ji} =
\begin{cases}
\bS_{ji} + \lambda_e e_j d_i + \lambda_{\text{imp}} p_i, & C_{ji}=1,\\
-\infty, & C_{ji}=0.
\end{cases}
\end{equation}
The candidate mask is applied after score fusion so excluded shots remain infeasible.
$\mathbf{e} \in \R^J$ is per-bar energy, $\mathbf{d} \in \R^I$ is per-shot visual dynamics ($\ell_2$ norm of consecutive feature differences), and $\mathbf{p} \in \R^I$ is shot importance.
The energy term increases the preference for dynamic shots as musical energy rises.
The shot selector produces the final sequence from $\bS^{\text{comb}}$ based on Algorithm~\ref{alg:bardp}.

\paragraph{Phase 4: Composition.}
A VLM-based Tagline Generator (Qwen3-VL-30B) watches a montage of key shots and produces short tagline texts with a recommended style; a Text Card Generator based on Qwen-Image~\citep{wu2025qwen} renders them as stylized overlays.
Each shot is trimmed to its bar duration with energy-driven transitions (dissolve / flash / hard cut).
Silero VAD~\citep{silero2021vad} detects dialogue for audio ducking, and all components are rendered into the final trailer via FFmpeg.

\paragraph{Phase 5: VLM Critic.}
Qwen3-VL-30B~\citep{bai2025qwen3vl} watches the rendered trailer at 1fps and rates it across five perceptual dimensions (1--10 each).
If quality is insufficient, the critic returns timestamped text feedback identifying problematic segments and reasons, which is mapped back to bar indices to ban corresponding shots and trigger re-selection and re-rendering for up to 3 rounds.

%% file: sections/benchmark.tex
\section{TrailerArena: A Benchmark for Trailer Generation}
\label{sec:benchmark}

Prior trailer generation methods primarily use set-based (e.g., F1) and ordering metrics (e.g., LD, AA)~\citep{wang2024inverse,zhu2025self}.
Index-based overlap can reward shots concentrated in a narrow movie region while overlooking visual similarity.
These metrics leave visual distributions, overall trajectory similarity, and perceptual appeal unassessed.
\textbf{TrailerArena} addresses these gaps with 20+ metrics across four complementary dimensions (Table~\ref{tab:dimensions}).

\input{tables/arena_dimensions}

\paragraph{A. Shot Selection.}
Set-based metrics (F1, F1@$K$, IoU) measure binary overlap with GT; soft metrics (SoftF1@$K$, Chamfer Distance) award partial credit for near-misses.
Neither directly measures \emph{visual} similarity.
Analogous to FID, \textbf{Fr\'{e}chet Shot Distance} (FSD) computes the Fr\'{e}chet distance between Gaussians fitted to ImageBind features of predicted and GT shot sets:
\begin{equation}
\label{eq:fsd}
\text{FSD} = \|\boldsymbol{\mu}_{\hat{\cS}} {-} \boldsymbol{\mu}_{\cS^*}\|^2 + \operatorname{Tr}\!\bigl(\boldsymbol{\Sigma}_{\hat{\cS}} {+} \boldsymbol{\Sigma}_{\cS^*} {-} 2(\boldsymbol{\Sigma}_{\hat{\cS}} \boldsymbol{\Sigma}_{\cS^*})^{1/2}\bigr).
\end{equation}
The mean and covariance terms capture mismatches in content \emph{type} and diversity, respectively, providing a method-agnostic measure of visual-semantic distributions.
Appendix~\ref{app:metrics} (Table~\ref{tab:fsd_scenarios}) compares FSD, F1, and Chamfer Distance through diagnostic scenarios.

\paragraph{B. Ordering.}
Levenshtein Distance (LD) and Alignment Accuracy (AA) evaluate shot ordering on the overlap subsequence, following the SSMP protocol~\cite{zhu2025self}.
LD penalizes both missing shots and wrong ordering; AA measures pairwise ordering agreement.
Kendall $\tau$ provides a complementary global ordering signal on the full predicted sequence.

\paragraph{C. Composition.}
Sequence DTW (SDTW) compares predicted and GT trajectories in visual feature space, jointly capturing selection, ordering, and visual similarity.
Aesthetic Quality (AQ) measures whether the method picks visually striking frames via a pretrained CLIP aesthetic predictor.

\paragraph{D. Perceptual.}
A vision-language model (Qwen3-VL-30B)~\cite{bai2025qwen3vl} watches each generated trailer and rates it on Pacing, Coherence, Variety, Engagement, and Narrative (1--10 scale).

%% file: tables/arena_dimensions.tex
\begin{table}[t]
\centering
\caption{TrailerArena evaluation dimensions, quality questions, and representative metrics. The complete metrics can be found in Appendix~\ref{app:metrics}, Table~\ref{tab:metric_inventory}.}
\vspace{-0.1in}
\label{tab:dimensions}
\small
\begin{tabular}{llcl}
\toprule
Dim. & Question & \# Metrics & Representative Metrics \\
\midrule
\textbf{\tikz\fill[metrics1] (0,0) circle (.6ex);\ \  Selection} & Did the method select the right shots? & 8 & F1,  SoftF1@$K$, FSD \\
\textbf{\tikz\fill[metrics2] (0,0) circle (.6ex);\ \  Ordering} & Are shots ranked and ordered correctly? & 8 & LD, AA, $\tau$, R@$K$, mAP@$t$ \\
\textbf{\tikz\fill[metrics3] (0,0) circle (.6ex);\ \  Composition} & Is the assembled trailer well-composed? & 3 & SDTW, AQ, BeatAlign \\
\textbf{\tikz\fill[metrics4] (0,0) circle (.6ex);\ \  Perceptual} & Does the trailer look/feel good to a viewer? & 6 & VL-Overall \\
\bottomrule
\end{tabular}
\vspace{-0.15in}
\end{table}

%% file: sections/experiments.tex
\input{tables/main_results}

\section{Experiments}
\label{sec:experiments}

\subsection{Implementation Details}

\paragraph{Dataset.}
For the training datasets,
Stage-1 pretraining uses $\sim$4,500 video-music pairs from V2M~\citep{tian2025vidmuse}.
Stage-2 fine-tuning uses 870 trailer-movie pairs with ground-truth bar-to-shot alignments: 472 from MMSC~\citep{zhu2025weakly} and 398 from CMTD~\citep{wang2024inverse}.
For the test datasets, following MMSC~\citep{zhu2025weakly}, we use Test-8 and Test-73 for fair evaluation.

\paragraph{Baselines.}
We compare our proposed method with the state-of-the-art trailer generation methods V2T~\citep{irie2010automatic}, M2T, PPBVAM~\citep{xu2015trailer}, IPOT~\citep{wang2024inverse}, MMSC~\citep{zhu2025weakly}, and SSMP~\citep{zhu2025self}.
When evaluating the capability of shot selection, we also include video summarization baselines VASNet~\citep{fajtl2018summarizing}, CLIP-It~\citep{narasimhan2021clip}, OTVS~\citep{wang2023self}, agentic editing model CutClaw~\citep{cutclaw2025}, and commercial software Muvee~\citep{ganhor2014muvee}.

\paragraph{Model architecture and Hyperparameter settings.}
We use ImageBind~\citep{girdhar2023imagebind} ($d_v{=}1024$) as the visual encoder and CLAP HTSAT-tiny~\citep{wu2023large} ($d_a{=}512$) at 48kHz as the acoustic encoder.
We use TransNetV2~\citep{soucek2024transnet} for shot detection and Ruptures~\citep{truong2020selective} for change-point detection.
All experiments are conducted on 1 NVIDIA H200 GPU.
We use Qwen-Image~\citep{wu2025qwen} for text-over-video shot generation.
Appendix~\ref{app:training-settings} provides the MuVA architecture and training hyperparameters.

\subsection{Quantitative Evaluation on TrailerArena}
\label{sec:main_results}

Table~\ref{tab:main_results} compares all four TrailerArena dimensions on Test-8 and Test-73.
BEAT leads shot selection (F1, SoftF1@5, FSD) on both sets.
SoftF1@5 additionally credits nearby shot indices, while the lowest FSD indicates the closest match to the GT visual feature distribution.
VASNet and OTVS achieve reasonable F1 but much higher FSD, favoring general visual saliency over trailer-appropriate content.
SSMP leads AA on both sets and LD on Test-73, while IPOT leads LD on Test-8.
BEAT ranks second on both LD and AA on Test-73.
Muvee leads $\tau$ on Test-8 by selecting few, long shots that preserve chronological order; BEAT ranks second, with RoPE-based self-attention capturing temporal structure.
BEAT also achieves the lowest SDTW on both sets, indicating the closest overall trajectory to GT in selection, ordering, and visual content.
It leads VLM-rated perceptual quality on both sets, with the largest gain on Test-8.
Table~\ref{tab:ablation2} further examines the effects of individual pipeline components and selection strategies.

\begin{figure}[t]
\setlength{\abovecaptionskip}{0pt}
    \centering
    \includegraphics[width=1.0\textwidth]{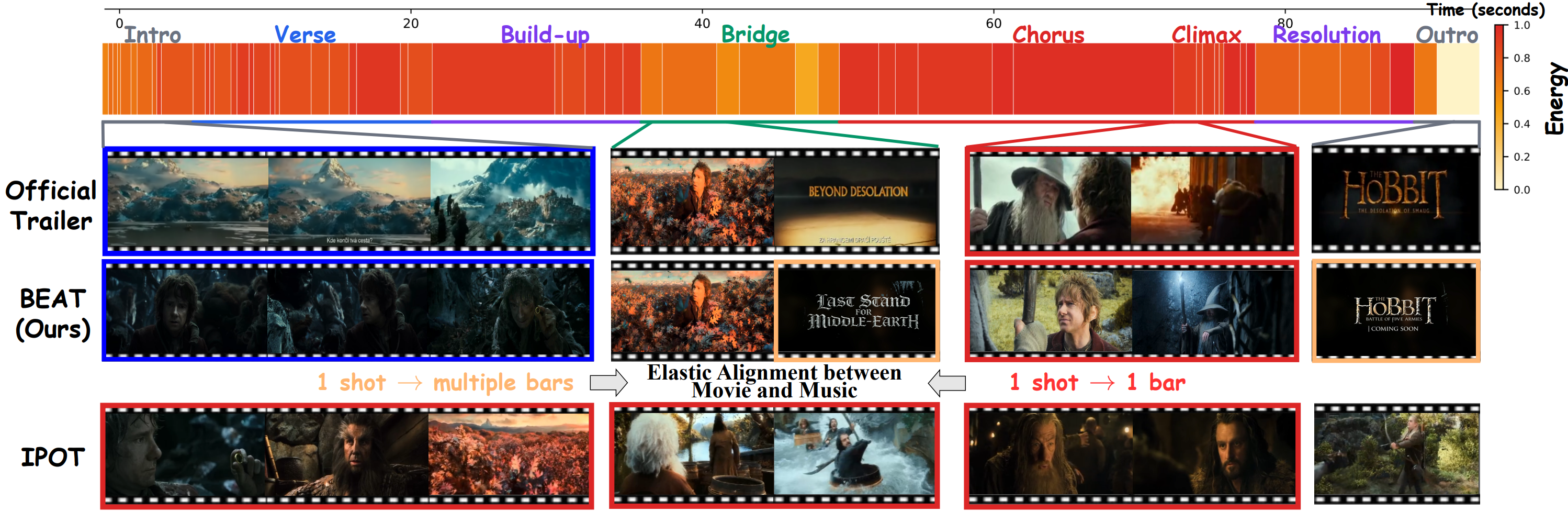}
    \caption{
    Comparison between the trailer generated by our method against the official trailer and IPOT~\citep{wang2024inverse}.
    The top row shows the breakdown of the trailer's music into bars, along with the energy level for each bar.
    The shots enclosed in {\color{blue}blue boxes} are long takes that span multiple music bars, while the {\color{red}red boxes} indicate quick cuts where each bar corresponds to a separate shot.
    The shots enclosed in {\color{yellow}gold boxes} are text-over-video shots generated by our pipeline using Qwen-Image~\citep{wu2025qwen}.
    }
    \label{fig:vis1}
\end{figure}

\subsection{Qualitative Analysis}
\label{sec:qualitative}

\begin{wrapfigure}[13]{r}{0.28\textwidth}
\centering
\includegraphics[width=\linewidth]{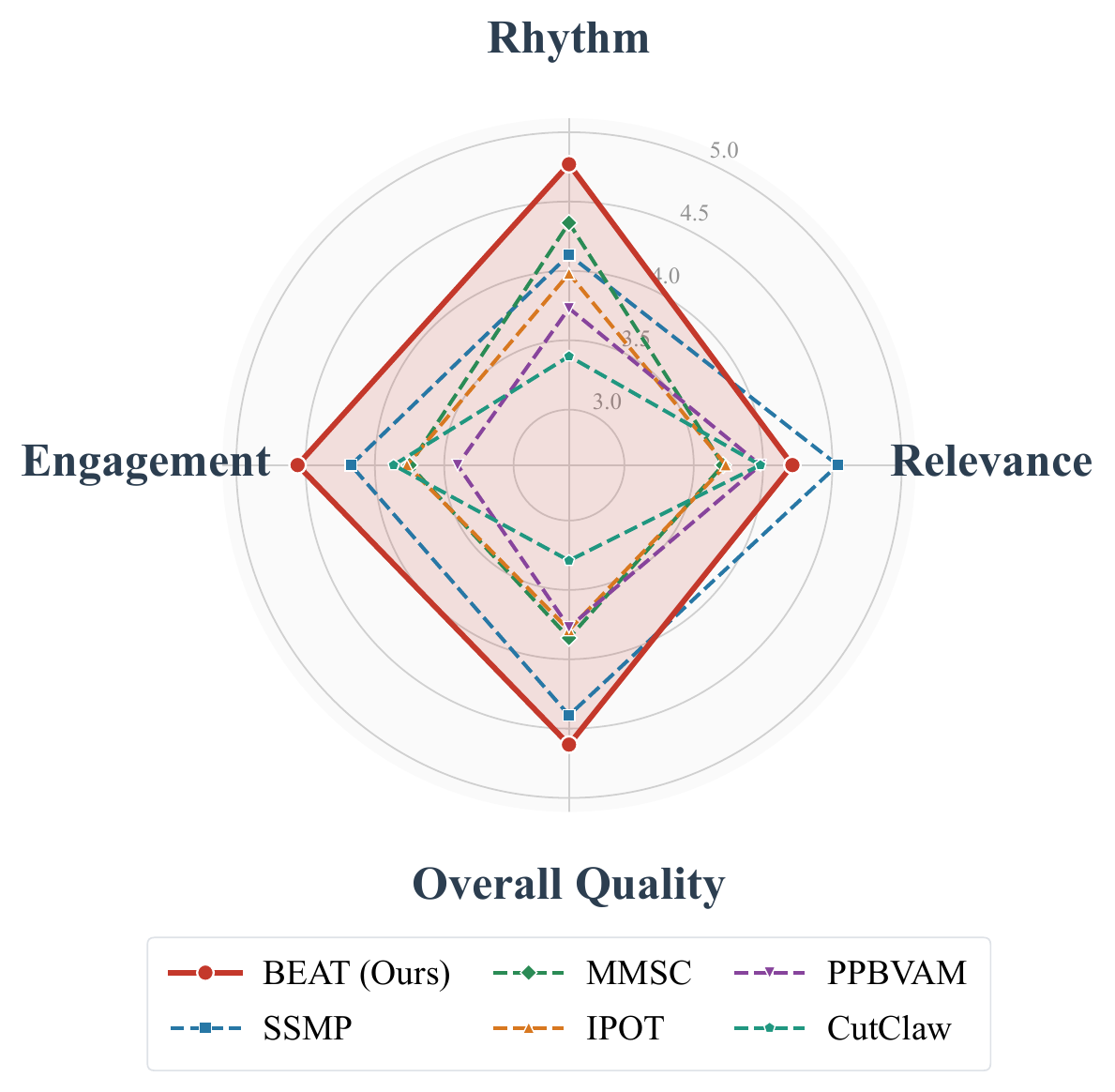}
\caption{User study.}
\label{fig:user_study}
\end{wrapfigure}
Figure~\ref{fig:vis1} compares BEAT and IPOT with the official trailer on one test movie.
BEAT adapts its editing rhythm to musical dynamics: it sustains shots across multiple bars during the low-energy bridge (blue boxes), mirroring the official trailer's pacing, and uses rapid one-bar-one-shot cuts during the high-energy chorus (red boxes).
IPOT instead assigns one shot per bar throughout, producing a uniform rhythm.
BEAT also integrates stylized text cards at low-energy moments (gold boxes) directly into its generation pipeline.
In our user study, 26 participants rate trailers from all methods on two Test-8 movies using a five-point Likert scale for rhythm alignment, shot relevance, engagement, and overall quality.
BEAT significantly outperforms all five baselines in rhythm and engagement, and MMSC, IPOT, PPBVAM, and CutClaw in overall quality; SSMP scores significantly higher on relevance.
Appendix~\ref{app:human} provides the study design, statistical analysis, and detailed results.

\input{tables/ablation_results}

\begin{figure*}[t]
\setlength{\abovecaptionskip}{0pt}
    \centering
    \includegraphics[width=1.0\textwidth]{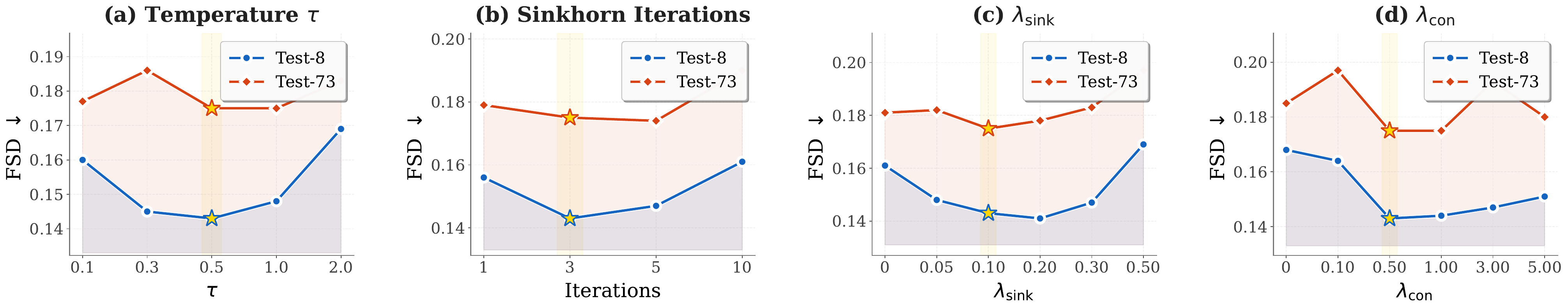}
    \caption{
    Ablation study of training hyperparameters. Stars mark defaults.
    }
    \label{fig:ablation}
\end{figure*}

\input{tables/ablation2_results}

\subsection{Ablation Study}
\label{sec:ablation}

\paragraph{MuVA training (Table~\ref{tab:ablation}).}
We ablate four design choices for training MuVA.
\textbf{(a)} CLAP outperforms ImageBind Audio and Mel-CLIP across all metrics, likely because its LAION-Audio-630K pretraining provides stronger music-specific representations than general-purpose audio encoders.
\textbf{(b)} Each loss term contributes incrementally: KL alignment provides the base supervision, InfoNCE anchors the cross-modal embedding space, and Sinkhorn regularization further improves both FSD and AA by preventing degenerate score concentration.
\textbf{(c)} Two-stage training achieves the best FSD.
Skipping V2M pretraining hurts distributional quality; skipping trailer fine-tuning yields poor AA because the model cannot specialize to trailer-specific ordering.
\textbf{(d)} Combining CMTD and MMSC (870 pairs) outperforms either alone: CMTD-only suffers from limited scale, while MMSC-only achieves competitive FSD but weaker ordering.

\paragraph{Pipeline modules (Table~\ref{tab:ablation2}).}
\textbf{(a)} Each agent shifts the quality profile in a predictable way.
MusicAnalyzer improves $\tau$ and VL by providing energy-aware bar segmentation that enables rhythm-matched editing.
VLM Critic trades F1 for higher VL by re-selecting shots for coherence.
NarrativeGuard reduces F1 because excluding late-movie shots narrows the candidate pool, but is essential for spoiler-free output.
\textbf{(b)} Text cards improve AQ but slightly reduce VL, while transitions show the opposite trade-off.
\textbf{(c)} Bar-DP trades F1 for improved VL, AA, and $\tau$: elastic multi-bar spans produce more perceptually appealing trailers, consistent with professional editing practice where long takes are preferred during low-energy passages.

%% file: tables/main_results.tex
\begin{table*}[t]
\centering
\caption{TrailerArena main results on Test-8 (8 movies) and Test-73 (73 movies). Best in \textbf{bold}, second best \underline{underlined}. Methods for video summarization are marked with a dagger ($\dagger$).}
\vspace{-0.1in}
\label{tab:main_results}
\small
\setlength{\tabcolsep}{3.5pt}
\resizebox{\linewidth}{!}{%
\begin{tabular}{cl ccc ccc cc c}
\toprule
& & \multicolumn{3}{c}{\tikz\fill[metrics1] (0,0) circle (.6ex);\ \  \textbf{Selection}} & \multicolumn{3}{c}{\tikz\fill[metrics2] (0,0) circle (.6ex);\ \  \textbf{Ordering}} & \multicolumn{2}{c}{\tikz\fill[metrics3] (0,0) circle (.6ex);\ \  \textbf{Composition}} &
\multicolumn{1}{c}{\tikz\fill[metrics4] (0,0) circle (.6ex);\ \  \textbf{Perceptual}}\\
\cmidrule(lr){3-5} \cmidrule(lr){6-8} \cmidrule(lr){9-10} \cmidrule(lr){11-11}
Test & Method
  & F1$\uparrow$ & SF1@5$\uparrow$ & FSD$\downarrow$
  & LD$\downarrow$ & AA$\uparrow$ & $\tau$$\uparrow$
  & SDTW$\downarrow$ & AQ$\uparrow$
  & VL-Overall$\uparrow$ \\
\midrule

\multirow{10}{*}{\rotatebox{90}{\normalsize \textbf{Test-8}}}
& VASNet$^{\dagger}$~\citep{fajtl2018summarizing}  &0.096  &0.276  &0.197     &100.6  &0.441  &-0.004  &0.351  &4.026  &-- \\
& CLIP-It$^{\dagger}$~\citep{narasimhan2021clip}  &0.074  &0.231  &0.247     &101.2  &0.414  &-0.025  &0.382  &4.026  &-- \\
& OTVS$^{\dagger}$~\citep{wang2023self}  &0.076  &0.243  &0.201     &101.4  &0.475  &-0.040  &0.368  &3.724  &-- \\
& Muvee$^{\dagger}$~\citep{ganhor2014muvee} &0.065  &0.129  &0.362     &103.5  &0.370  &\textbf{0.271}  &0.396  &4.255  &5.53 \\
& CutClaw~\citep{cutclaw2025}      &0.036  &0.292  &0.433  &100.3  &0.027  &0.034  &0.352  &4.147  &6.13 \\
& V2T~\citep{irie2010automatic}  &0.054  &0.154  &0.240     &103.8  &0.529  &0.025  &0.404  &4.288  &5.85 \\
& PPBVAM~\citep{xu2015trailer}  &0.083  &0.213  &\underline{0.191}     &101.5  &0.531  &0.040  &0.373  &4.288  &5.25 \\
& IPOT~\citep{wang2024inverse}  & 0.131 & 0.278 &0.259     & \textbf{99.8} & \underline{0.538} &0.055  &\underline{0.349}  &3.721  &\underline{6.48} \\
& MMSC~\citep{zhu2025weakly}  & 0.139 & 0.271 &0.238     & 101.1 &0.394  & $-$0.017 &0.392  &3.724  &6.30 \\
& SSMP~\citep{zhu2025self}      & \underline{0.162} & \underline{0.324} & 0.198 & \underline{100.0} & \textbf{0.568} & $-$0.036 &0.390  &\underline{4.856}  &6.15 \\
& \cellcolor{tabhighlight}\textbf{BEAT (Ours)}          & \cellcolor{tabhighlight}\textbf{0.193} & \cellcolor{tabhighlight}\textbf{0.373} & \cellcolor{tabhighlight}\textbf{0.143} & \cellcolor{tabhighlight}100.2 & \cellcolor{tabhighlight}0.529 & \cellcolor{tabhighlight}\underline{0.060} &\cellcolor{tabhighlight}\textbf{0.330}  &\cellcolor{tabhighlight}\textbf{4.919}  & \cellcolor{tabhighlight}\textbf{6.58}\\
\midrule

\multirow{7}{*}{\rotatebox{90}{\normalsize \textbf{Test-73}}}
& VASNet$^{\dagger}$~\citep{fajtl2018summarizing}  &0.077  &0.235  &0.260     &84.8  &0.441  &0.003  &0.367  &4.182  &-- \\
& CLIP-It$^{\dagger}$~\citep{narasimhan2021clip}  &0.046  &0.173  &0.316   &85.8  &0.389  &0.013  &0.371  &4.096  &-- \\
& OTVS$^{\dagger}$~\citep{wang2023self}  &0.077  &0.235  &0.276     &84.8  &0.400  &-0.012  &0.375  &4.237  &-- \\
& IPOT~\citep{wang2024inverse}  & 0.104 & 0.262 & 0.241    &83.6  &0.505  &0.028  &0.365  &\textbf{4.291}  &\underline{5.86} \\
& MMSC~\citep{zhu2025weakly}  &0.197  &0.280  &0.222     &82.5  &0.500  & 0.020 &0.367  &4.197  &5.81 \\
& SSMP~\citep{zhu2025self}      & \underline{0.226} & \underline{0.325} &\underline{0.212}   &\textbf{77.3}  &\textbf{0.526}  & \textbf{0.095} &\underline{0.333}  &4.262  &5.81 \\
& \cellcolor{tabhighlight}\textbf{BEAT (Ours)}          & \cellcolor{tabhighlight}\textbf{0.233} & \cellcolor{tabhighlight}\textbf{0.403} & \cellcolor{tabhighlight}\textbf{0.175} &\cellcolor{tabhighlight}\underline{81.1}  &\cellcolor{tabhighlight}\underline{0.523}  & \cellcolor{tabhighlight}\underline{0.049} &\cellcolor{tabhighlight}\textbf{0.323}  &\cellcolor{tabhighlight}\underline{4.263}  &\cellcolor{tabhighlight}\textbf{5.94} \\
\bottomrule
\end{tabular}%
}
\vspace{-0.15in}
\end{table*}

%% file: tables/ablation_results.tex
\begin{table*}[ht]
    \centering
    \caption{\textbf{Ablation study} for MuVA on Test-73.  
    Default settings are marked as \colorbox[HTML]{E6E6E6}{gray}.}
    \tabcolsep=3.0pt
    \begin{subtable}[t]{0.45\linewidth}
    \centering
        \small
        \begin{tabular}{ccccc}
            \toprule
            Arch & Objective  & FSD$\downarrow$ & AA$\uparrow$ & SDTW$\downarrow$ \\ 
            \midrule
            SwinT& CLAP   &\cellcolor[HTML]{E6E6E6}\textbf{0.175} &\cellcolor[HTML]{E6E6E6}\textbf{0.523} &\cellcolor[HTML]{E6E6E6}\textbf{0.323} \\
            ViT-H/14& ImageBind   &0.186 &0.490 &0.352 \\
            ViT-B/32& Mel-CLIP     &0.181 &0.511 &0.334 \\
            \bottomrule
        \end{tabular}
        \caption{\textbf{Audio encoders.}}
    \end{subtable}
    \hspace{0.1cm}
    \begin{subtable}[t]{0.45\linewidth}
    \centering
        \small
        \begin{tabular}{cccccc}
            \toprule
            $\mathcal{L}_{\text{KL}}$ & $\mathcal{L}_{\text{InfoNCE}}$ & $\mathcal{L}_{\text{Sink}}$ & FSD$\downarrow$ & AA$\uparrow$ & SDTW$\downarrow$ \\ 
            \midrule
            \cmark &\xmark &\xmark &0.191 &0.506 &0.331 \\
            \cmark &\cmark &\xmark &0.181 &0.519 &0.329 \\
            \cmark &\cmark &\cmark &\cellcolor[HTML]{E6E6E6}\textbf{0.175} &\cellcolor[HTML]{E6E6E6}\textbf{0.523} &\cellcolor[HTML]{E6E6E6}\textbf{0.323}  \\
            \bottomrule
        \end{tabular}
        \caption{\textbf{Loss terms} for Stage 2. 
        }
    \end{subtable}
    \\ 
    \begin{subtable}[t]{0.48\linewidth}
    \centering
        \small
        \begin{tabular}{ccccc}
            \toprule
            MV & Movie-Trailer & FSD$\downarrow$ & AA$\uparrow$ & SDTW$\downarrow$ \\ 
            \midrule
            \cmark &\xmark  &0.227 &0.415 &\textbf{0.312}  \\ 
            \cmark &\cmark &\cellcolor[HTML]{E6E6E6}\textbf{0.175} &\cellcolor[HTML]{E6E6E6}0.523 &\cellcolor[HTML]{E6E6E6}0.323 \\
            \xmark &\cmark  &0.184 &\textbf{0.539} &0.333 \\
            \bottomrule
        \end{tabular}
        \caption{\textbf{Training stages.}}
    \end{subtable}
    \hspace{0.1cm}
    \begin{subtable}[t]{0.48\linewidth}
    \centering
        \small
        \begin{tabular}{cccccc}
            \toprule
            CMTD &MMSC  & \#Pairs & FSD$\downarrow$ & AA$\uparrow$ & SDTW$\downarrow$\\ 
            \midrule
            \cmark &\xmark &398 &0.217 &0.452 &0.347  \\ 
            \xmark &\cmark &472 &0.178 &0.517 &0.324  \\ 
            \cmark &\cmark &870 &\cellcolor[HTML]{E6E6E6}\textbf{0.175} &\cellcolor[HTML]{E6E6E6}\textbf{0.523} &\cellcolor[HTML]{E6E6E6}\textbf{0.323}  \\ 
            \bottomrule
        \end{tabular}
        \caption{\textbf{Training datasets} for Stage 2.}
    \end{subtable}
    
    \label{tab:ablation}
    \vspace{-0.3cm}
\end{table*}

%% file: tables/ablation2_results.tex
\begin{table*}[ht]
    \centering
    \caption{\textbf{Ablation study} for the module of BEAT on Test-73.}
    \tabcolsep=3pt
    \begin{subtable}[t]{0.52\linewidth}
    \centering
        \small
        \begin{tabular}{lcccc}
        \toprule
        Configuration & F1$\uparrow$  & AA$\uparrow$  & $\tau$$\uparrow$ & VL-Overall$\uparrow$ \\
        \midrule
        MuVA  &0.233 &0.523 &0.049 &5.94 \\
        $+$ MusicAnalyzer &0.231   &0.532  &0.057  &6.02   \\
        $+$ VLM Critic &0.215  &0.511  &0.065  &6.14  \\
        $+$ NarrativeGuard &0.195  &0.528  &0.040  &5.79 \\
        \bottomrule
        \end{tabular}
        \caption{\textbf{Role of each module.}}
    \end{subtable}
    \hspace{0.1cm}
    \begin{subtable}[t]{0.43\linewidth}
    \centering
        \small
        \begin{tabular}{lcc}
        \toprule
        Configuration & AQ$\uparrow$  & VL-Overall$\uparrow$ \\
        \midrule
        MuVA &4.263 &5.94  \\
        $+$ Text Cards &4.310 &5.90  \\
        $+$ Transition &4.211 &6.03  \\
        \bottomrule
        \end{tabular}
        \caption{\textbf{Trailer Composition}. 
        }
    \end{subtable}
    \\ 
    \begin{subtable}[t]{\linewidth}
    \centering
        \small
        \begin{tabular}{lccccccccccc}
        \toprule
        Strategy & & & F1$\uparrow$ & SF1@5$\uparrow$ & FSD$\downarrow$
          & LD$\downarrow$ & AA$\uparrow$ & $\tau$$\uparrow$
          & SDTW$\downarrow$ & AQ$\uparrow$
          & VL-Overall$\uparrow$ \\
        \midrule
        Greedy & & &0.233 &0.403  &0.175 &81.1 &0.523  &0.049  &0.323  &4.263 &5.94  \\
        Bar-DP & & &0.217 &0.378  &0.180 &82.7 &0.530  &0.054  &0.327  &4.346 &6.23   \\
        \bottomrule
        \end{tabular}
        \caption{\textbf{Shot selection strategy.}}
    \end{subtable}
    \label{tab:ablation2}
\end{table*}

%% file: sections/conclusion.tex
\section{Conclusion}
\label{sec:conclusion}


We presented BEAT, a music-guided movie trailer generation framework that combines learned bar-shot compatibility, energy-adaptive shot durations, and agent-based creative planning. MuVA and Bar-DP provide explicit control over shot selection and timing, while structured text signals coordinate planning, composition, and iterative critique. The resulting pipeline takes a full-length movie and a chosen soundtrack as input and produces a fully composed trailer. We also introduced TrailerArena for evaluation across four complementary dimensions, with experiments demonstrating leading shot-selection performance on both test sets. BEAT illustrates how trainable audiovisual alignment and explicit editing constraints can support flexible agentic workflows for trailer production.

%% file: sections/appendix.tex
\appendix
\raggedbottom

\input{sections/appendix_contents}

\input{sections/appendix_evaluation}
\input{sections/appendix_results}

\input{sections/appendix_human_study}
\input{sections/appendix_discussion}

%% file: sections/appendix_contents.tex
\begin{center}
  {\Large\bfseries Appendix\par}
\end{center}
\vspace{0.65em}

\begingroup
\hypersetup{hidelinks}
\definecolor{appendixred}{HTML}{D62728}
\setlength{\fboxsep}{8pt}
\setlength{\fboxrule}{0.4pt}
\noindent\fcolorbox{black}{white}{%
\begin{minipage}{\dimexpr\linewidth-2\fboxsep-2\fboxrule\relax}
\raggedright
{\normalsize\scshape Contents\par}
\vspace{0.45em}
\small
\setlength{\parskip}{0.12em}
\newcommand{\appsectionentry}[2]{%
  \noindent{\color{appendixred}\hyperref[#1]{%
    \makebox[2.2em][l]{\textbf{\ref*{#1}}}\textbf{\MakeUppercase{#2}}}}%
  \hfill\textbf{\pageref{#1}}\par
}
\newcommand{\appsubsectionentry}[2]{%
  {\footnotesize\noindent\hspace*{1.2em}%
  {\color{appendixred}\hyperref[#1]{%
    \makebox[3.2em][l]{\ref*{#1}}\MakeUppercase{#2}}}%
  \nobreak\leaders\hbox to 0.55em{\hss.\hss}\hfill\nobreak
  \pageref{#1}\par}%
}

\appsectionentry{app:evaluation}{Experimental Protocol and Metrics}
\appsubsectionentry{app:task-data}{Task and Evaluation Data}
\appsubsectionentry{app:training-settings}{MuVA Architecture and Training Hyperparameters}
\appsubsectionentry{app:bardp-settings}{Bar-DP Hyperparameters}
\appsubsectionentry{app:configurations}{System Configurations}
\appsubsectionentry{app:metrics}{Evaluation Metrics}
\appsectionentry{app:additional_results}{Additional Quantitative Results}
\appsubsectionentry{app:retrieval}{Standalone Music-Bar-to-Shot Retrieval}
\appsubsectionentry{app:selection}{Shot-Selection Precision and Recall}
\appsubsectionentry{app:rhythm}{Rhythmic Synchronization}
\appsubsectionentry{app:runtime}{Computational Cost}
\appsectionentry{app:human}{Human Evaluation}
\appsubsectionentry{app:human-design}{Study Design and Participants}
\appsubsectionentry{app:human-analysis}{Statistical Analysis}
\appsubsectionentry{app:human-results}{Results}
\appsectionentry{app:discussion}{Discussion and Limitations}
\end{minipage}%
}
\endgroup
\vspace{0.75em}

%% file: sections/appendix_evaluation.tex
\section{Experimental Protocol and Metrics}
\label{app:evaluation}

\subsection{Task and Evaluation Data}
\label{app:task-data}
BEAT addresses music-guided editing: given a source movie and a selected soundtrack,
it selects and times shots according to the music's semantics, structure, and dynamics.
This setting assumes that music is available before visual editing, while professional
workflows may also select or revise music during editing. It complements methods
that generate or retrieve music for fixed visuals.

We evaluate on the established Test-8 and Test-73 splits using TrailerArena's metrics
and protocols, which also apply to other generated and ground-truth (GT) trailer
collections. Reproducing the results requires access to the source movies and
trailers; copyright restrictions and unavailable media links may limit this access.

\subsection{MuVA Architecture and Training Hyperparameters}
\label{app:training-settings}
MuVA uses a latent dimension of $d{=}512$, $L{=}6$ Transformer blocks,
$L_{\text{ca}}{=}3$ cross-attention layers, and 8 attention heads.
Stage 1 pretraining runs for 50 epochs with a batch size of 64, an initial learning
rate of $3{\times}10^{-4}$, and a cosine learning-rate schedule with warm-up.
The optimizer parameters are $\beta_1{=}0.9$ and $\beta_2{=}0.999$.
Stage 2 fine-tuning uses movie-trailer pairs for 20 epochs, with a batch size of 16
and a learning rate of $5{\times}10^{-5}$. We set the InfoNCE and Sinkhorn weights
to $\lambda_{\text{con}}{=}0.5$ and $\lambda_{\text{sink}}{=}0.1$, respectively,
and use Sinkhorn temperature $\tau_s{=}0.5$ with $T{=}3$ normalization iterations.

\subsection{Bar-DP Hyperparameters}
\label{app:bardp-settings}
For Bar-DP (Section~\ref{sec:bardp}), we use beam width $W{=}50$, retain the top
$M{=}20$ candidate shots per bar, and allow spans of up to $k_{\max}{=}5$ bars.
The smoothness and cut-bonus weights are $\lambda_{\text{sm}}{=}0.3$ and
$\lambda_{\text{cut}}{=}0.5$, respectively.
The duration constraint uses $\eta{=}0.9$, allowing a shot to span up to
$1/\eta\approx1.11$ times its original duration through mild slow-motion.
The visual similarity threshold for neighbor exclusion is $\theta_{\text{sim}}{=}0.80$.
The single-CPU-core runtime reported in Section~\ref{sec:bardp} uses these settings
with approximately $J{=}60$ music bars and $I{=}1500$ movie shots.

\subsection{System Configurations}
\label{app:configurations}
The main comparison in Table~\ref{tab:main_results} evaluates MuVA with greedy,
rigid one-to-one bar-to-shot selection. The VLM Critic and generated text cards are
disabled. Table~\ref{tab:ablation2} separately evaluates elastic selection with
Bar-DP and the optional agentic components. These comparisons separate the effects of the alignment model,
selection strategy, and optional composition modules.

When enabled, the critic inspects a 1-fps representation for semantic and narrative
issues, such as irrelevant or incoherent segments. MuVA and Bar-DP perform
fine-grained alignment using audio- and shot-level representations. This sampling rate supports high-level inspection but may miss brief
cuts and transition artifacts. Appendix~\ref{app:rhythm} therefore evaluates
rhythmic synchronization separately.

\subsection{Evaluation Metrics}
\label{app:metrics}
Table~\ref{tab:metric_inventory} expands the four dimensions in
Table~\ref{tab:dimensions} into 25 entries (8 selection, 8 ordering, 3 composition,
and 6 perceptual). For this inventory, F1@$K$, SoftF1@$K$, and mAP@$t$ each count
as one parameterized metric family; R@1, R@5, and R@10 are listed separately.
The perceptual count includes the five individual ratings and their mean.
Additional threshold variants and diagnostic scores do not add entries to this count.

\input{tables/appendix_metric_inventory}

The ordering dimension includes both sequence-order and per-bar retrieval metrics.
The latter require a bar-by-shot score matrix; their protocol is described in
Appendix~\ref{app:retrieval}. The evaluation script's mAP@$t$ uses the reciprocal
rank of the first shot within $t$ indices of the GT-aligned shot, averaged over bars,
with $t\in\{0,1,2,3\}$. It is thus a tolerance-based reciprocal-rank score,
rather than conventional average precision over all relevant shots.
SoftF1 is evaluated at $K\in\{1,3,5\}$, with $K{=}5$ in the main comparison.
BeatAlign uses a $\pm0.2$\,s window around music-bar boundaries; separate detected-event
rhythm diagnostics are reported in Appendix~\ref{app:rhythm}.

Seven of the nine metrics in Table~\ref{tab:main_results} use the GT trailer:
F1, SoftF1@5, FSD, LD, AA, Kendall $\tau$, and SDTW. AQ and VL-Overall complement
these reference-based measurements with visual-attractiveness and model-based
perceptual scores.

\paragraph{Shot-set precision and recall.}
For predicted and GT shot sets $\hat{\mathcal S}$ and $\mathcal S^*$, respectively,
\begin{equation}
 P=\frac{|\hat{\mathcal S}\cap\mathcal S^*|}{|\hat{\mathcal S}|},\qquad
 R=\frac{|\hat{\mathcal S}\cap\mathcal S^*|}{|\mathcal S^*|},\qquad
 F1=\frac{2PR}{P+R}.
\end{equation}
Precision measures the fraction of selected shots present in GT; recall measures
the fraction of GT shots recovered. Appendix~\ref{app:selection} reports both
quantities alongside the F1 summary used in the main table.

\paragraph{Comparison of shot-selection metrics.}
Table~\ref{tab:fsd_scenarios} contrasts the information captured by F1, Chamfer
Distance, and FSD. FSD operates on visual features rather than shot indices.
Its mean and covariance terms characterize shifts in content and feature variation,
complementing exact-overlap and index-proximity measurements.
For the synthetic examples, GT shots have indices $\{10,20\}$ and scalar features
$\{0,2\}$. Shots 9 and 19 have features 0 and 2, respectively, while shots 11 and
21 both have feature 0. The illustrative Chamfer distance is the average of the
two directional mean nearest-neighbor distances, using absolute index differences.
Thus, $\{9,19\}$ and $\{11,21\}$ have identical F1 and index-distance scores, but
only the former matches the GT feature distribution. Identical shot sets yield
F1${=}1$ and FSD${=}0$; FSD does not distinguish distributions with identical
means and covariances.

\input{tables/appendix_fsd_scenarios}

\paragraph{Aesthetic Quality (AQ).}
In the selection-only setting, we sample the midpoint frame of each unique selected
source-movie shot, extract a CLIP ViT-L/14 representation, and apply a pretrained
aesthetic MLP. Trailer-level AQ is the mean frame score. It measures visual attractiveness;
ordering, pacing, and music synchronization require separate temporal metrics.

\paragraph{VL-Overall.}
In the main comparison, VLM ratings are computed after generation and provide no
feedback to the system. The critic-enabled ablation uses related VLM judgments
during refinement, making VL-Overall partly dependent on the refinement criterion.
Appendix~\ref{app:discussion} discusses this dependence, and
Appendix~\ref{app:human} provides independent human ratings.

%% file: tables/appendix_metric_inventory.tex
\begin{table}[t]
\centering
\caption{Full metric inventory for Table~\ref{tab:dimensions}. Arrows indicate the preferred direction; GT indicates whether a reference trailer or its annotations are required.}
\label{tab:metric_inventory}
\small
\setlength{\tabcolsep}{3pt}
\renewcommand{\arraystretch}{1.12}
\begin{tabular}{@{}p{0.17\linewidth}p{0.23\linewidth}cp{0.49\linewidth}@{}}
\toprule
Dimension & Metric & GT & Measurement \\
\midrule
\multirow{8}{*}{Selection (8)}
 & Precision $\uparrow$ & Yes & Fraction of selected shots present in GT. \\
 & Recall $\uparrow$ & Yes & Fraction of GT shots recovered. \\
 & F1 $\uparrow$ & Yes & Harmonic mean of shot-set precision and recall. \\
 & F1@$K$ $\uparrow$ & Yes & F1 with hard shot-index tolerance $K$. \\
 & IoU $\uparrow$ & Yes & Intersection over union of selected and GT shot sets. \\
 & SoftF1@$K$ $\uparrow$ & Yes & F1 with linearly decreasing credit for nearby indices. \\
 & Chamfer distance $\downarrow$ & Yes & Symmetric mean nearest-neighbor shot-index distance. \\
 & FSD $\downarrow$ & Yes & Distance between selected and GT visual feature distributions. \\
\midrule
\multirow{8}{*}{Ordering (8)}
 & LD $\downarrow$ & Yes & Edit distance under the overlap-subsequence protocol. \\
 & AA $\uparrow$ & Yes & Pairwise order agreement among overlapping shots. \\
 & Kendall $\tau$ $\uparrow$ & Yes & Rank correlation between predicted and GT sequences. \\
 & R@1 $\uparrow$ & Yes & Per-bar GT-shot retrieval success at rank 1. \\
 & R@5 $\uparrow$ & Yes & Per-bar GT-shot retrieval success within the top 5. \\
 & R@10 $\uparrow$ & Yes & Per-bar GT-shot retrieval success within the top 10. \\
 & MRR $\uparrow$ & Yes & Mean reciprocal rank of the GT-aligned shot. \\
 & mAP@$t$ $\uparrow$ & Yes & Threshold-tolerant per-bar ranking score (see text). \\
\midrule
\multirow{3}{*}{Composition (3)}
 & SDTW $\downarrow$ & Yes & DTW distance between visual feature trajectories. \\
 & AQ $\uparrow$ & No & Mean pretrained aesthetic score of selected-shot frames. \\
 & BeatAlign $\uparrow$ & No & Fraction of music-bar boundaries near a shot transition. \\
\midrule
\multirow{6}{*}{Perceptual (6)}
 & VL-Pacing $\uparrow$ & No & VLM rating of editing pace. \\
 & VL-Coherence $\uparrow$ & No & VLM rating of coherence across shots. \\
 & VL-Variety $\uparrow$ & No & VLM rating of visual variety. \\
 & VL-Engagement $\uparrow$ & No & VLM rating of viewer engagement. \\
 & VL-Narrative $\uparrow$ & No & VLM rating of narrative quality. \\
 & VL-Overall $\uparrow$ & No & Mean of the five perceptual ratings. \\
\bottomrule
\end{tabular}
\end{table}

%% file: tables/appendix_fsd_scenarios.tex
\definecolor{rowgray}{gray}{0.93}
\begin{table}[H]
\centering
\caption{Shot-selection metrics: properties and controlled synthetic examples. The examples illustrate index overlap, index proximity, and visual-distribution agreement; they are not experimental results.}
\label{tab:fsd_scenarios}
\small
\setlength{\tabcolsep}{3pt}
\begin{tabular}{@{}l lll@{}}
\toprule
 & \textbf{F1}$\uparrow$ & \textbf{Chamfer Dist.}$\downarrow$ & \textbf{FSD}$\downarrow$ \\
\midrule
\textit{Space}        & Shot indices & Shot indices & Visual features \\
\rowcolor{rowgray}
\textit{Granularity}  & Binary (hit/miss) & NN distance & Distribution ($\mu$+$\Sigma$) \\
\textit{Measures}     & Index overlap & Index proximity & Content + diversity \\
\midrule
\multicolumn{4}{@{}l}{\textit{Synthetic examples: GT indices $\{10,20\}$, features $\{0,2\}$}} \\[3pt]
\rowcolor{rowgray}
Identical GT shots: $\{10,20\}$ & 1 & 0 & 0 \\
Matching features: $\{9,19\}$ & 0 & 1 & 0 \\
\rowcolor{rowgray}
Collapsed features: $\{11,21\}$ & 0 & 1 & $>0$ \\
Half GT, matching features: $\{10,19\}$ & 0.5 & 0.5 & 0 \\
\bottomrule
\end{tabular}
\end{table}

%% file: sections/appendix_results.tex
\section{Additional Quantitative Results}
\label{app:additional_results}

We evaluate local bar-shot retrieval, trailer-level shot selection, and the
rhythmic structure of rendered trailers, then report selection and composition
runtimes separately.

\input{sections/appendix_retrieval}
\input{sections/appendix_selection}
\input{sections/appendix_rhythm}
\input{sections/appendix_runtime}

%% file: sections/appendix_retrieval.tex
\subsection{Standalone Music-Bar-to-Shot Retrieval}
\label{app:retrieval}

To isolate the alignment model, we evaluate MuVA's bar-by-shot compatibility matrix
on Test-73 without Bar-DP, agents, the critic, or the compositor. Each of the 6,415
valid music bars is a query over all shots in its source movie, with an average
of approximately 1,381 candidates per movie.

Exact R@$K$ records whether the annotated GT-aligned shot appears among the top
$K$ candidates. MRR averages the reciprocal rank $1/r$ of that shot. We average query scores within each trailer, then average over all 73 trailers
to give each trailer equal weight. A temporally tolerant R@10 also
accepts either immediately adjacent shot ($\pm1$ shot), accounting for ambiguity
near editing boundaries. These query-level retrieval metrics evaluate a different
property from the shot-set recall defined in Appendix~\ref{app:evaluation}.

\input{tables/appendix_retrieval}

Table~\ref{tab:app_retrieval} shows that MuVA improves every reported retrieval
metric over frozen ImageBind cosine similarity and IPOT. Exact R@10 increases from
0.0510 for frozen ImageBind to 0.1177 for MuVA, while MRR increases from 0.0284 to
0.0635. These gains support learning bar-shot compatibility, although absolute recall
remains modest in the large candidate pools. Retrieval evaluates local matching;
selection and composition must additionally account for trailer-level constraints.

%% file: tables/appendix_retrieval.tex
\begin{table}[H]
\centering
\caption{Standalone bar-to-shot retrieval on Test-73. All values are proportions; higher is better. Dashes denote unreported values.}
\label{tab:app_retrieval}
\small
\setlength{\tabcolsep}{4pt}
\begin{tabular}{lrrrrr}
\toprule
& \multicolumn{4}{c}{Exact match} & $\pm1$ shot \\
\cmidrule(lr){2-5}
Method & R@1 & R@5 & R@10 & MRR & R@10 \\
\midrule
Random ranking & 0.0007 & 0.0036 & 0.0072 & -- & -- \\
IPOT & 0.0019 & 0.0087 & 0.0139 & 0.0085 & 0.0296 \\
Frozen ImageBind cosine & 0.0141 & 0.0361 & 0.0510 & 0.0284 & 0.0732 \\
MuVA & \textbf{0.0318} & \textbf{0.0879} & \textbf{0.1177} & \textbf{0.0635} & \textbf{0.1505} \\
\bottomrule
\end{tabular}
\end{table}

%% file: sections/appendix_selection.tex
\subsection{Shot-Selection Precision and Recall}
\label{app:selection}

We next examine the shot sets produced by the main-comparison configuration.
Table~\ref{tab:app_pr} expands the F1 results in Table~\ref{tab:main_results} into
precision and recall for IPOT, MMSC, SSMP, and BEAT on both test splits.

\input{tables/appendix_precision_recall}

BEAT achieves the highest precision and recall among these four methods on both
splits. The F1 gains reflect both more accurate selections and greater GT-shot coverage.
These trailer-level set metrics complement per-bar retrieval, but neither measures
shot timing relative to music. The next analysis evaluates this temporal aspect.

%% file: tables/appendix_precision_recall.tex
\begin{table}[H]
\centering
\caption{Exact shot-set precision, recall, and F1 on both test splits. Higher is better. F1 values match Table~\ref{tab:main_results}.}
\label{tab:app_pr}
\small
\begin{tabular}{llrrr}
\toprule
Split & Method & Precision & Recall & F1 \\
\midrule
Test-8 & IPOT & 0.122 & 0.143 & 0.131 \\
& MMSC & 0.130 & 0.150 & 0.139 \\
& SSMP & 0.156 & 0.168 & 0.162 \\
& BEAT & \textbf{0.180} & \textbf{0.208} & \textbf{0.193} \\
\midrule
Test-73 & IPOT & 0.097 & 0.113 & 0.104 \\
& MMSC & 0.184 & 0.214 & 0.197 \\
& SSMP & 0.211 & 0.243 & 0.226 \\
& BEAT & \textbf{0.217} & \textbf{0.252} & \textbf{0.233} \\
\bottomrule
\end{tabular}
\end{table}

%% file: sections/appendix_rhythm.tex
\subsection{Rhythmic Synchronization}
\label{app:rhythm}

The rigid one-to-one configuration places cuts at predefined music-bar boundaries,
giving IPOT, MMSC, SSMP, and BEAT the same BeatAlign score of 1.0. To assess
rhythmic structure beyond the imposed bar grid, we evaluate Test-73 using
the official All-In-One beat/downbeat tracker~\citep{kim2023allinone} and
independently detected visual cuts.

Reference-free evaluation measures whether cuts occur near tracked beats or
downbeats. Reference-based evaluation measures whether generated and GT trailers
cut near the same tracked music events. We report beat and downbeat F1 separately,
using $\pm70$\,ms as the primary tolerance and $\pm200$\,ms for sensitivity analysis.

\input{tables/appendix_rhythm}

The event-based scores distinguish methods that bar-grid BeatAlign treats equally
(Table~\ref{tab:app_rhythm}). MuVA-Greedy achieves the highest beat and downbeat F1
in both evaluation settings at both tolerances, ahead of MuVA-BarDP and BEAT-Full.
Sustained shots reduce cut density and may trade musical-event coverage for visual
continuity; elastic pacing therefore does not uniformly improve beat matching.
The human rhythm ratings in Appendix~\ref{app:human} complement these event-level
scores by assessing the perceived quality of the editing rhythm.

%% file: tables/appendix_rhythm.tex
\begin{table}[H]
\centering
\caption{Rhythmic synchronization on Test-73 using tracked music events and detected visual cuts. All columns report F1; higher is better. Reference-free scores measure cut-to-music alignment, whereas reference-based scores measure agreement with the GT rhythmic editing pattern.}
\label{tab:app_rhythm}
\small
\begin{tabular}{lrrrr}
\toprule
& \multicolumn{2}{c}{Reference-free} & \multicolumn{2}{c}{Reference-based} \\
\cmidrule(lr){2-3}\cmidrule(lr){4-5}
Method & Beat & Downbeat & Beat & Downbeat \\
\midrule
\multicolumn{5}{l}{\textit{Primary tolerance: $\pm70$\,ms}} \\
IPOT & 0.101 & 0.054 & 0.095 & 0.118 \\
MMSC & 0.104 & 0.052 & 0.080 & 0.066 \\
SSMP & 0.104 & 0.050 & 0.085 & 0.103 \\
MuVA-Greedy & \textbf{0.128} & \textbf{0.059} & \textbf{0.099} & \textbf{0.124} \\
MuVA-BarDP & 0.084 & 0.045 & 0.065 & 0.082 \\
BEAT-Full & 0.065 & 0.055 & 0.073 & 0.093 \\
\midrule
\multicolumn{5}{l}{\textit{Sensitivity analysis: $\pm200$\,ms}} \\
IPOT & 0.277 & 0.151 & 0.245 & 0.284 \\
MMSC & 0.299 & 0.151 & 0.232 & 0.265 \\
SSMP & 0.281 & 0.143 & 0.223 & 0.249 \\
MuVA-Greedy & \textbf{0.345} & \textbf{0.169} & \textbf{0.265} & \textbf{0.303} \\
MuVA-BarDP & 0.213 & 0.127 & 0.180 & 0.198 \\
BEAT-Full & 0.183 & 0.145 & 0.166 & 0.210 \\
\bottomrule
\end{tabular}
\end{table}

%% file: sections/appendix_runtime.tex
\subsection{Computational Cost}
\label{app:runtime}

We measure selection time with cached features on one NVIDIA H200 and report
median runtimes in Table~\ref{tab:app_runtime}. MuVA-Greedy takes 0.0094\,s and
MuVA with Bar-DP takes 0.297\,s, compared with 0.0005\,s for IPOT and 0.0045\,s
for MMSC. These timings exclude feature extraction and cover selection rather
than complete video generation.

\input{tables/appendix_runtime}

The reported median runtime of the agentic composition pipeline is 115.3\,s,
dominated by VLM inference, text-card generation, and video rendering. These
operations cost substantially more than shot-index selection. The separate timing
scopes quantify this overhead but do not establish an end-to-end speed advantage
over the baselines.

%% file: tables/appendix_runtime.tex
\begin{table}[H]
\centering
\caption{Median runtime in seconds. Selection uses cached features on one NVIDIA H200. Agentic composition is timed separately and covers a broader set of operations.}
\label{tab:app_runtime}
\small
\begin{tabular}{lr}
\toprule
Method / timed scope & Median time (s) \\
\midrule
\multicolumn{2}{l}{\textit{Selection only, cached features}} \\
IPOT & 0.0005 \\
MMSC & 0.0045 \\
MuVA-Greedy & 0.0094 \\
MuVA + Bar-DP & 0.297 \\
\midrule
Agentic composition pipeline & 115.3 \\
\bottomrule
\end{tabular}
\end{table}

%% file: sections/appendix_human_study.tex
\section{Human Evaluation}
\label{app:human}

The user study complements automatic measurements of shot overlap, alignment,
and rhythmic events with viewer assessments of the composed trailers. We report
the design, participant-level analysis, and results for the original and expanded
cohorts.

\subsection{Study Design and Participants}
\label{app:human-design}
Following the user-study setting of MMSC, we use two Test-8 movies:
\emph{300: Rise of an Empire} and \emph{The Hobbit: The Desolation of Smaug}.
Participants view trailers generated by BEAT, MMSC, PPBVAM, SSMP, IPOT, and CutClaw,
with each movie's official trailer provided as a quality reference. Generated
trailers are anonymized, and the method order differs between the two movies to
reduce order bias. Every participant rates every generated trailer on a five-point
Likert scale using four questions:
\begin{itemize}
\item \textbf{Rhythm:} How well do the visual cuts align with the rhythm of the music?
\item \textbf{Engagement:} How engaging is the trailer?
\item \textbf{Relevance:} How well does the trailer convey the theme and style of the source movie?
\item \textbf{Overall quality:} How would you rate the trailer's overall quality, including how closely it resembles a real movie trailer?
\end{itemize}

The study initially included 12 participants, primarily aged 20--30, and was
expanded with 14 additional participants: ten aged 20--30, two aged 30--40, and
two aged 40--50. Section~\ref{sec:qualitative} reports the full 26-participant
cohort; the two cohorts are nested, not independent.

\subsection{Statistical Analysis}
\label{app:human-analysis}
For each participant and method, we average the two movie scores. We apply a
repeated-measures Friedman test across the six methods, followed by two-sided paired
Wilcoxon signed-rank tests with Holm correction across the five BEAT--baseline
comparisons within each metric and cohort ($\alpha=0.05$). Zero differences are
excluded; we use exhaustive sign permutations for $N=12$ and a tie-adjusted normal
approximation without continuity correction for $N=26$.
The omnibus test indicates differences among methods; superiority to individual
baselines is assessed by the corrected pairwise tests.

\subsection{Results}
\label{app:human-results}
For the original 12 participants, pairwise comparisons favor BEAT primarily in
rhythm, engagement, and overall quality, with no significant advantage in
relevance. In the expanded cohort, BEAT receives significantly higher rhythm and engagement
ratings than all five baselines, and higher overall-quality ratings than MMSC,
IPOT, PPBVAM, and CutClaw.

For relevance, BEAT scores significantly above MMSC and IPOT but below SSMP;
differences from PPBVAM and CutClaw are not significant. These results support
perceptual gains on specific dimensions for the two evaluated movies. Limited
movie coverage and participant diversity constrain their generalizability.

%% file: sections/appendix_discussion.tex
\section{Discussion and Limitations}
\label{app:discussion}

Rhythmic alignment and perceived editing quality capture complementary aspects of
trailer generation. MuVA-Greedy achieves the highest tracked-event F1, while human
ratings favor BEAT's rhythm and engagement on the evaluated movies. Frequent cuts
can increase musical-event coverage, whereas sustained shots can preserve visual
continuity. These results motivate evaluating elastic editing with both temporal
metrics and viewer judgments.

The optional modules also affect evaluation dimensions differently
(Table~\ref{tab:ablation2}). Text cards raise AQ from 4.263 to 4.310 but lower
VL-Overall from 5.94 to 5.90; the critic raises VL-Overall from 5.94 to 6.14 but
reduces GT F1 from 0.233 to 0.215. Neither yields uniform gains. Because the critic
and evaluator use related VLM judgments, the perceptual-score increase may partly
reflect agreement with the refinement criterion. A dedicated human comparison of
critic-enabled and critic-disabled outputs is needed to isolate its benefit.

Several limitations remain. Alignment inherits potential biases from ImageBind
and CLAP and may favor salient visual or acoustic content over narratively
important but subtle scenes. Generated text cards and transitions may contain
artifacts, and selected shots may form incoherent sequences. The critic's limited
temporal resolution can leave brief failures undetected.

The two-movie human study and qualitative examples provide limited coverage of
possible outputs, while the scale of public movie-trailer datasets constrains
broader evaluation. Generalization to documentaries, animation, and non-Western
cinema remains underexplored. Evaluating more genres with broader participant
populations, while holding movie and music inputs fixed across methods, would
strengthen the evidence. Applications to music videos, promotional montages, and
highlight reels also remain untested.

Practical deployment depends on access to source media and the VLM and
image-generation components. Tagline generation and critique add inference cost,
and model inference and rendering dominate composition time. Changes in component
availability or behavior can also affect reproducibility. Fast shot selection
therefore does not by itself imply low-cost trailer production.